\documentclass[runningheads]{llncs}
\usepackage[T1]{fontenc}
\usepackage{graphicx}

\usepackage{amssymb}

\usepackage{longtable}
\usepackage{booktabs}
\usepackage{listings}
 \usepackage{multirow}
\usepackage{makecell}

\usepackage{booktabs}
\usepackage{tikz}
\usepackage{graphicx}
\usepackage{tcolorbox}

\newcommand{\AllSmall}[3][0.5]{
\begin{tikzpicture}[scale=#1]
    \draw (0,0) circle (2cm);
    \node at (0.3,1.6) {\scriptsize #3};
    \draw (-0.3,-0.2) circle (1.2cm);
    \node at (-0.2,0.5) {\scriptsize #2};
\end{tikzpicture}
}

\newcommand{\NoSmall}[3][0.5]{
\begin{tikzpicture}[scale=#1]
    \draw (0,0) circle (1.2cm);
    \node at (0, 0.7) {\scriptsize #2};
    \draw (3.0,0) circle (1.2cm);
    \node at (3.0, 0.7) {\scriptsize #3};
\end{tikzpicture}
}

\newcommand{\SomeSmall}[3][0.5]{
\begin{tikzpicture}[scale=#1]
    \draw (-0.8,0) circle (1.5cm);
    \node at (-1.1, 1.0) {\scriptsize #2};
    \draw (0.8,0,0) circle (1.5cm);
    \node at (1.1, 1.0) {\scriptsize #3};
    \node at (0,0) {$\times$};
\end{tikzpicture}
}

\newcommand{\SomeNotSmall}[3][0.5]{
\begin{tikzpicture}[scale=#1]
    \draw (-0.8,0) circle (1.5cm);
    \node at (-1.1, 1.0) {\scriptsize #2};
    \draw (0.8,0,0) circle (1.5cm);
    \node at (1.1, 1.0) {\scriptsize #3};
    \node at (-1.4,-0.4) {$\times$};
\end{tikzpicture}
}

\newcommand{\myparagraph}[1]{
\medskip
\noindent\textbf{#1}
}

\begin{document}
\title{Do Diagrams Help Large Language Models Reason?
Evidence from Syllogistic Reasoning}
\titlerunning{Do Diagrams Help Large Language Models Reason?}
\author{Risako Ando \and Koji Mineshima}
\authorrunning{R. Ando and K. Mineshima}
\institute{Keio University, 2-15-45 Mita, Minato-ku, Tokyo 108-8345, Japan}
\maketitle 
\begin{abstract}

Diagrams are widely used to support logical reasoning, and prior studies suggest that representations such as Euler diagrams can improve human reasoning performance. Recent work has also explored their effects on large language models (LLMs). In this paper, we compare four representational conditions for syllogistic reasoning: natural language, logical notation, linear diagrams, and Euler diagrams. Using 285 problems from Ando et al.~(2024), we evaluate two contemporary LLMs, Claude~3.5~Sonnet and GPT-4o-mini.
Our results show that diagrammatic representations do not consistently improve performance. Although the models perform well on entailment and contradiction problems, they struggle with neutral problems and often make systematic conversion errors. Overall, the results suggest that the tested models gain limited benefit from diagrams in logical reasoning tasks.


\end{abstract}

\section{Introduction}
\label{sec:introduction}

Diagrams have long been used as external representations to support logical reasoning~\cite{stenning2001diagrammatic,Shimojima2015}. In diagrammatic reasoning research, visual representations such as Euler diagrams can make set-theoretic relations spatially explicit and thereby facilitate deductive reasoning. A number of empirical studies have demonstrated that diagrammatic representations can improve human performance on syllogistic reasoning tasks. In particular, Euler diagrams have been shown to support deductive reasoning effectively and may outperform traditional Venn diagrams \cite{sato-mineshima-takemura2010,sato2015diagrams}. Similarly, linear diagrams have been shown to provide reasoning support comparable to Euler diagrams in syllogistic tasks \cite{sato2012efficacy,sato2015diagrams}.

Recently, large language models (LLMs) have achieved strong performance on a wide range of reasoning tasks. However, it remains unclear whether diagrammatic representations can support reasoning in LLMs in the same way as they do for human reasoners.
Ando et al.~\cite{ando2024can} recently investigated this question by examining whether Euler diagrams improve syllogistic reasoning in LLMs. Their experiments suggested that providing Euler diagrams alongside textual premises can improve LLM performance.

In this paper, we revisit this question by systematically comparing different representational conditions for syllogistic reasoning. 
While Ando et al.~\cite{ando2024can} compared natural language and Euler diagram representations, the present study extends this setup by including two additional representational conditions: linear diagrams and symbolic logical forms. This enables a more comprehensive comparison across textual, symbolic, and diagrammatic representations within a unified experimental framework.

Using a dataset of 285 syllogistic problems from the validity checking task introduced in~\cite{ando2024can}, we evaluate two contemporary multimodal LLMs, Claude~3.5~Sonnet and GPT-4o-mini. The dataset includes annotations that enable us to analyze several well-known reasoning phenomena, including conversion errors, figural effects, and differences across inference labels and content types.

Our results show that differences across representational conditions are relatively small, and that diagrammatic representations do not consistently improve reasoning accuracy.
In addition, the models exhibit substantial conversion errors and biases across inference labels, performing poorly on neutral cases.
Overall, the findings suggest that the multimodal LLMs examined in this study gain limited benefit from diagrammatic representations for logical inference.

\begin{table}[t]
\centering

\caption{Four representational conditions used in syllogistic reasoning.}

\begin{tabular}{llll}  \toprule
\textbf{Natural language} \hspace{2ex} &
\textbf{Logical form} \hspace{2ex} &
\textbf{Euler diagram} \hspace{2ex} &
\textbf{Linear diagram} \\ \midrule

\textsf{All $S$ are $P$} &
$S \subseteq P$ &
\begin{minipage}{4em}
\AllSmall[0.25]{S}{P}
\end{minipage} &
\begin{minipage}{9em}
\ttfamily
------- S\\
-------------- P
\end{minipage} \\

\addlinespace[1mm] 
\midrule

\textsf{No $S$ are $P$} &
$S \cap P = \emptyset$ &
\begin{minipage}{4em}
\medskip
\NoSmall[0.3]{S}{P}
\end{minipage} &
\begin{minipage}{9em}
\ttfamily
----- S\\
\hspace*{1cm}------ P
\end{minipage}
\\
\midrule

\addlinespace[1mm] 

\textsf{Some $S$ are $P$} &
$S \cap P \neq \emptyset$ &
\begin{minipage}{4em}
\SomeSmall[0.3]{S}{P}
\end{minipage} &
\begin{minipage}{10em}
\ttfamily
-----x--- S\\
\hspace*{1.2ex}----x------- P
\end{minipage} \\

\midrule

\textsf{Some $S$ are not $P$} \hspace{1ex} &
$S \setminus P \neq \emptyset$ &
\begin{minipage}{4em}
\smallskip
\SomeNotSmall[0.3]{S}{P}
\end{minipage} &
\begin{minipage}{10em}
\ttfamily
---x--------- S\\
\hspace*{1cm}------ P
\end{minipage} \\

\bottomrule
\end{tabular}
\label{tab:categorical}
\end{table}

\section{Background}

Syllogistic reasoning is a classical form of deductive reasoning involving two premises and a conclusion that relate three terms. A typical example is: \textit{All $B$ are $C$; Some $A$ are $B$; Therefore, some $A$ are $C$}.
The quantified sentence forms used in syllogisms are summarized in Table~\ref{tab:categorical}.
Because syllogisms provide a well-controlled setting for studying logical inference, they have been widely used in cognitive psychology~\cite{geurts2003reasoning}.

Diagrammatic representations provide an alternative way to express logical relations between sets.
As shown in Table~\ref{tab:categorical}, Euler diagrams represent sets as regions in the plane, where spatial inclusion and exclusion represent subset and disjoint relations.
Such diagrams make set-theoretic relations visually explicit and have been studied as tools for supporting logical reasoning
\cite{stenning2001diagrammatic,Shimojima2015}.
Empirical studies have shown that Euler diagrams can facilitate human syllogistic reasoning and may outperform Venn diagrams in syllogistic reasoning tasks~\cite{sato-mineshima-takemura2010,sato2015diagrams}.

Another form of diagrammatic representation is the \emph{linear diagram}, in which sets are represented by horizontal line segments (see Table~\ref{tab:categorical}).
Inclusion and exclusion relations are encoded by the relative positions of the segments. In both Euler and linear diagrams, an ``x'' denotes the existence of at least one element.
Experimental studies have shown that linear diagrams can support human reasoning performance at a level comparable to Euler diagrams \cite{sato2012efficacy,sato2015diagrams}.

Recent multimodal LLMs can process both text and images, making it possible to investigate whether diagrammatic representations affect reasoning performance \cite{liu2023llava,openai2023gpt4v}. 
Ando et al.~\cite{ando2024can} reported that Euler diagrams can improve performance on certain reasoning patterns, such as the neutral and conversion cases discussed in Section~\ref{sec:setup}.
In this work, we systematically compare textual, symbolic, and diagrammatic representations within a unified experimental framework.

\section{Experimental Setup}
\label{sec:setup}

\subsection{Dataset}

We evaluate the models on a dataset of 285 syllogistic reasoning problems derived from the validity checking (VC) task introduced by Ando et al.~\cite{ando2024can}. Each problem consists of two premises and a conclusion relating three terms, and the task is to determine the logical relation between the premises and the conclusion.
Table~\ref{tab:examples} shows representative examples together with their annotations.

Following standard treatments of syllogistic reasoning, each problem is assigned one of three inference labels: \textit{entailment}, \textit{contradiction}, or \textit{neutral}. An \textit{entailment} label indicates that the conclusion logically follows from the premises, a \textit{contradiction} label indicates that the premises and the conclusion cannot both be true, and a \textit{neutral} label indicates that the conclusion is neither entailed nor contradicted by the premises.

The dataset also includes problems annotated as \emph{conversion} cases.
Conversion errors arise from incorrectly reversing quantified statements, such as interpreting \textit{All A are B} as \textit{All B are A}, or \textit{Some A are not B} as \textit{Some B are not A} (see \cite{sato2015diagrams}). In our dataset, these problems are labeled as \emph{neutral}, although such incorrect conversions would make the conclusions appear entailed.
For example, in the first problem shown in Table~\ref{tab:examples}, incorrectly converting \textit{All B are A} into \textit{All A are B} would make the conclusion appear entailed, although the correct inference label is \emph{neutral}.

Syllogisms are also classified according to their \emph{figure}, namely the arrangement of the middle term across the premises. Different figures are known to vary in difficulty for human reasoners, a phenomenon known as the \emph{figural effect}, with Figure 4 typically being more difficult than Figure 1 (see~\cite{geurts2003reasoning,sato2015diagrams}).
Each problem is therefore annotated with one of the four syllogistic figures.

In addition, the natural language formulations are categorized into three content types: \emph{symbolic}, \emph{congruent}, and \emph{incongruent} (see Table~\ref{tab:examples}). Symbolic problems use abstract symbols (e.g., $A$, $B$, $C$), whereas congruent and incongruent problems differ in whether the described relations are semantically plausible. This distinction allows us to examine whether model performance is influenced by semantic plausibility in addition to logical structure.

The dataset is publicly available online\footnote{\url{https://github.com/kmineshima/euler-diagrams-llm}}. 
All annotations used in this study, including inference labels, conversion cases, syllogistic figures, and content types, follow Ando et al.~\cite{ando2024can}; see their work for further details on the annotation scheme.

\begin{table*}[t]
\centering
\caption{Representative examples from the dataset with their annotations.}
\begin{tabular}{lcccc}
\toprule
\textbf{Example} & \textbf{Inference} & \textbf{Conversion} & \textbf{Figure} & \textbf{Content} \\
\midrule

\makecell[l]{
\textbf{P1:} All B are A. \\
\textbf{P2:} No B are C. \\
\textbf{C:} No C are A.
}
& Neutral & Yes & 3 & Symbolic \\
\midrule

\makecell[l]{
\textbf{P1:} All beer are liquor. \\
\textbf{P2:} Some drink are not liquor. \\
\textbf{C:} Some drink are not beer.
}
& Entailment & No & 2 & Congruent \\
\midrule

\makecell[l]{
\textbf{P1:} All animals are dogs. \\
\textbf{P2:} All robots are animals. \\
\textbf{C:} Some robots are not dogs.
}
& Contradiction & No & 1 & Incongruent \\

\bottomrule
\end{tabular}
\label{tab:examples}
\end{table*}

\subsection{Representational Conditions}

Each reasoning problem is presented under one of four representational conditions shown in Table~\ref{tab:categorical}. 
In all conditions, the premises and conclusion are presented in natural language, while additional symbolic or diagrammatic representations are provided depending on the condition.

\begin{description}
\item[Natural Language (NL)] Only the natural language statements are shown.
\item[Logic] Each premise is accompanied by a set-theoretic logical representation.
\item[Linear] Each premise is accompanied by a text-based (ASCII) linear diagram.
\item[Euler] Each premise is accompanied by an Euler diagram presented as an image.
\end{description}
The Linear and Euler diagram representations follow
Sato and Mineshima~\cite{sato2012efficacy,sato2015diagrams}.

\subsection{Models and Evaluation}

We evaluate two multimodal LLMs:
\textbf{Claude~3.5~Sonnet} and \textbf{GPT-4o-mini}, accessed through the OpenRouter API, representing widely used contemporary multimodal LLMs of different scales.

For each problem, the model is asked to determine the logical relation between the premises and the conclusion by selecting one of three labels: \emph{entailment}, \emph{contradiction}, or \emph{neutral}.
The models are instructed to respond using only the corresponding label.

Performance is measured by accuracy over the 285 problems.
In addition to overall accuracy, we analyze results across several dimensions: representational conditions (NL, Logic, Linear, Euler), conversion vs.\ non-conversion cases, syllogistic figure (Figures~1--4), inference label (entailment, contradiction, neutral), and content type (symbolic, congruent, incongruent).

The prompt format follows Ando et al.~\cite{ando2024can}, with additional instructions explaining the linear and Euler diagram representations.
Figure~\ref{fig:prompt-example} shows an example input for the Linear condition.

\begin{figure}[t]
\centering
\begin{minipage}{0.62\linewidth}
\begin{tcolorbox}[title={Example input (Linear condition)}]

\ttfamily

Premise 1: No A are B. \\
Linear diagram for Premise 1:

A:\hspace{1em}------- \\
B:\hspace{5em}-------

\smallskip

Premise 2: Some B are C. \\
Linear diagram for Premise 2:

B:\hspace{1em}-----x---- \\
C:\hspace{2em}--x---------

\smallskip

Conclusion: All C are A.

\end{tcolorbox}
\end{minipage}

\caption{
Example input in the Linear condition
(correct inference label: \textit{contradiction}).
}
\label{fig:prompt-example}
\end{figure}

\begin{table}[b]
\centering
\caption{Overall accuracy and accuracy on conversion vs.\ non-conversion cases (\%).}
\footnotesize
\begin{tabular}{l|ccc|ccc}
\hline
 & \multicolumn{3}{c|}{Claude 3.5 Sonnet} & \multicolumn{3}{c}{GPT-4o-mini} \\
Condition & Overall & Non-conv & Conv & Overall & Non-conv & Conv \\
\hline
NL     & 84.6 & 84.6 & 84.2 & 68.8 & 69.9 & 52.6 \\
Logic  & 84.6 & 84.6 & 84.2 & 71.2 & 72.9 & 47.4 \\
Linear & 81.1 & 82.0 & 68.4 & 64.9 & 66.5 & 42.1 \\
Euler  & 84.2 & 85.0 & 73.7 & 67.0 & 69.2 & 36.8 \\
\hline
\end{tabular}
\label{tab:overall-conversion}
\end{table}

\section{Results}

Table~\ref{tab:overall-conversion} reports overall accuracy as well as accuracy on conversion and non-conversion cases across the four representational conditions.

Claude~3.5~Sonnet achieves relatively high performance across all conditions, with slightly lower accuracy in the Linear condition, whereas GPT-4o-mini performs substantially worse overall (65--71\%).
For both models, the Logic condition yields the highest accuracy, whereas the diagrammatic conditions fail to produce consistent improvements over NL and sometimes yield lower accuracy.

Overall, these results suggest that LLM performance varies only modestly across representational conditions, and that diagrammatic representations do not consistently improve reasoning accuracy, in contrast to findings from human reasoning studies.

Regarding conversion problems, Claude~3.5~Sonnet shows only small differences between conversion and non-conversion cases in the NL and Logic conditions, whereas performance decreases under the diagrammatic conditions.
GPT-4o-mini exhibits substantially larger gaps across all conditions, indicating stronger susceptibility to conversion-related reasoning errors.

Table~\ref{tab:figure-accuracy} presents accuracy by syllogistic figure.
Both models show systematic variation across figures, with Figure~1 generally yielding the highest accuracy and Figure~4 the lowest, consistent with the findings in human reasoning~\cite{sato2015diagrams}.

Table~\ref{tab:label-accuracy} reports accuracy across inference labels.
For both models, entailment and contradiction cases are solved with high accuracy, whereas neutral cases are more difficult.
This pattern is particularly pronounced for GPT-4o-mini, with accuracy on neutral cases below 30\% across all conditions.

\begin{table}[t]
\centering
\footnotesize
\caption{Accuracy by syllogistic figure (\%).}
\begin{tabular}{l|cccc|cccc}
\hline
 & \multicolumn{4}{c|}{Claude 3.5 Sonnet} & \multicolumn{4}{c}{GPT-4o-mini} \\
Condition & Figure 1 & Figure 2 & Figure 3 & Figure 4 & Figure 1 & Figure 2 & Figure 3 & Figure 4 \\
\hline
NL     & 91.3 & 80.0 & 89.8 & 82.7 & 89.1 & 66.7 & 69.5 & 58.7 \\
Logic  & 93.5 & 82.9 & 88.1 & 78.7 & 82.6 & 70.5 & 78.0 & 60.0 \\
Linear & 91.3 & 79.0 & 89.8 & 70.7 & 73.9 & 66.7 & 74.6 & 49.3 \\
Euler  & 91.3 & 81.0 & 94.9 & 76.0 & 87.0 & 64.8 & 66.1 & 58.7 \\
\hline
\end{tabular}
\label{tab:figure-accuracy}
\end{table}

\begin{table}[t]
\centering
\caption{Accuracy by inference label (\%).}
\footnotesize
\begin{tabular}{l|ccc|ccc}
\hline
 & \multicolumn{3}{c|}{Claude 3.5 Sonnet} & \multicolumn{3}{c}{GPT-4o-mini} \\
Condition & Entailment & Contradiction & Neutral & Entailment & Contradiction & Neutral \\
\hline
NL     & 97.9 & 92.6 & 63.2 & 81.1 & 100.0 & 25.3 \\
Logic  & 98.9 & 92.6 & 62.1 & 86.3 & 100.0 & 27.4 \\
Linear & 97.9 & 96.8 & 48.4 & 78.9 & 96.8 & 18.9 \\
Euler  & 95.8 & 100.0 & 56.8 & 82.1 & 98.9 & 20.0 \\
\hline
\end{tabular}
\label{tab:label-accuracy}
\end{table}

\begin{table}[t!]
\centering
\caption{Accuracy by content type (\%).}
\footnotesize
\begin{tabular}{l|ccc|ccc}
\hline
 & \multicolumn{3}{c|}{Claude 3.5 Sonnet} & \multicolumn{3}{c}{GPT-4o-mini} \\
Condition & Symbolic & Congruent & Incongruent & Symbolic & Congruent & Incongruent \\
\hline
NL     & 95.6 & 85.8 & 79.2 & 82.2 & 75.8 & 56.7 \\
Logic  & 86.7 & 87.5 & 80.8 & 73.3 & 74.2 & 67.5 \\
Linear & 86.7 & 83.3 & 76.7 & 68.9 & 68.3 & 60.0 \\
Euler  & 91.1 & 85.0 & 80.8 & 75.6 & 72.5 & 58.3 \\
\hline
\end{tabular}
\label{tab:content-type-accuracy}
\end{table}

We further analyze model performance across the three content types: \emph{symbolic}, \emph{congruent}, and \emph{incongruent}.
Table~\ref{tab:content-type-accuracy} summarizes accuracy across content types and representational conditions for both models.

Across models and conditions, performance generally decreases from symbolic to congruent and further to incongruent cases.
For example, in the NL condition, Claude~3.5~Sonnet achieves 95.6\% accuracy on symbolic problems, 85.8\% on congruent problems, and 79.2\% on incongruent problems. GPT-4o-mini shows a stronger drop, from 82.2\% on symbolic problems to 56.7\% on incongruent ones.

Overall, these results suggest that model performance is sensitive to semantic plausibility. Problems with semantically implausible content (incongruent cases) are substantially more difficult for both models, and diagrammatic representations do not mitigate this decrease in performance.

\section{Discussion}

Our results provide several insights into the reasoning behavior of LLMs across different representational conditions.

\myparagraph{Do Diagrams Help LLM Reasoning?}
A central question of this study was whether diagrammatic representations facilitate reasoning in LLMs in a manner similar to human cognition. In human reasoning, diagrams such as Euler diagrams or linear diagrams can externalize set relations and thereby support logical inference~\cite{sato-mineshima-takemura2010,sato2012efficacy,sato2015diagrams}.

However, our results indicate that diagrammatic representations provide limited benefit for LLMs. Across both models, reasoning accuracy remains broadly similar across natural language, logical form, and diagrammatic conditions. In particular, neither linear diagrams nor Euler diagrams substantially improve performance compared to textual representations.

Our results also differ somewhat from those reported by Ando et al.~\cite{ando2024can}, who found that Euler diagrams can improve LLM performance on the same validity-checking (VC) task. Since the experimental setup of the VC task is largely identical, one plausible explanation lies in differences between the models used in the two studies. This suggests that the effectiveness of diagrammatic representations for LLM reasoning may be highly model-dependent.

\myparagraph{Reasoning Biases in LLMs.}
Although performance differences across formats are small, the models exhibit several systematic reasoning patterns.

First, both models show lower accuracy on conversion problems than on non-conversion problems (Table~\ref{tab:overall-conversion}). The effect is particularly pronounced for GPT-4o-mini, but is also visible for Claude~3.5~Sonnet in the diagrammatic conditions.
In human reasoning, diagrams are known to help prevent such errors by making the difference between statements such as \textit{All $A$ are $B$} and \textit{All $B$ are $A$} visually explicit.
In contrast, we do not observe a comparable benefit for LLMs; for example, accuracy on conversion problems decreases under the Linear and Euler conditions for Claude~3.5~Sonnet, while GPT-4o-mini shows even larger conversion-related performance gaps in the diagrammatic conditions.

Second, both models show substantially lower accuracy on neutral problems than on entailment or contradiction (Table~\ref{tab:label-accuracy}).
Identifying neutral cases requires recognizing that a conclusion is neither logically implied nor incompatible with the premises, which appears to be considerably more difficult for current LLMs.

Third, performance decreases from symbolic to congruent and further to incongruent problems across both models and all conditions. Diagrammatic representations do not improve accuracy on incongruent cases, suggesting that model predictions depend more on semantic plausibility than on the structural relations expressed in the diagrams.

\myparagraph{Implications for Diagrammatic Reasoning in LLMs}
Taken together, our findings suggest that diagrammatic representations do not support LLM reasoning in the same way that they support human reasoning. While diagrams help humans by making set relations spatially explicit, LLMs appear able to extract sufficient relational information from textual or symbolic inputs alone.

More broadly, the results highlight a difference between human and machine reasoning. While diagrams support logical inference based on structural relations for humans, the LLMs we examined appear to rely more strongly on statistical patterns learned from textual data than on the spatial structure of diagrams.

\section{Conclusion}

This paper investigated whether diagrammatic representations improve syllogistic reasoning in LLMs.
Using the validity-checking dataset introduced by Ando et al.~\cite{ando2024can}, we compared four representational conditions: natural language, logical form, linear diagrams, and Euler diagrams.

Our results show that diagrammatic representations provide little advantage for current LLMs.
Across two models, performance varies only modestly across representational conditions, with the diagrammatic conditions, particularly the Linear condition, yielding lower accuracy than NL or Logic.
This contrasts with previous studies in human reasoning, where diagrams such as Euler diagrams and linear diagrams have been shown to facilitate syllogistic inference.

Our analysis also reveals systematic reasoning patterns in LLMs, including conversion errors and substantial difficulty with neutral inferences. 

Future work may explore whether multimodal training or diagram-aware architectures could enable models to make better use of diagrammatic information.

\myparagraph{Acknowledgment}
We thank the anonymous reviewers for their helpful comments and suggestions, which have improved this paper. We also thank Kentaro Ozeki for providing valuable comments. This work was partially supported by JST CREST Grant Number JPMJCR2114, JST BOOST Grant Number JPMJBS2409 and JSPS KAKENHI Grant Number JP24K00004.

\bibliographystyle{splncs04}
\bibliography{mybibliography}

\end{document}